\documentclass[11pt]{article}
\usepackage{authblk}
\usepackage{booktabs}
\usepackage{amssymb,amsmath,amsthm,multirow}
\usepackage{amsfonts,graphicx}
\usepackage{tikz}
\usepackage{float}
\usepackage[toc,page]{appendix}
\usepackage{caption}
\usepackage{subcaption}
\usepackage{algorithm}
\usepackage{algpseudocode}
\usepackage{comment}
\usepackage{indentfirst} 
\usepackage{url}

\newtheorem{theorem}{Theorem}

\newtheorem{lemma}{Lemma}

\newtheorem{dfn}{Definition}

\begin{document}
\title{Scalable unsupervised feature selection via weight stability}
\author{Xudong Zhang}
\author{Renato Cordeiro de Amorim\thanks{Corresponding author, \texttt{r.amorim@essex.ac.uk}}}
\affil{\small School of Computer Science and Electronic Engineering, University of Essex, Wivenhoe, UK.}
\date{}
\maketitle

\begin{abstract}
Unsupervised feature selection is critical for improving clustering performance in high-dimensional data, where irrelevant features can obscure meaningful structure. In this work, we propose the Minkowski weighted $k$-means++, a novel initialisation strategy for the Minkowski Weighted $k$-means. Our initialisation selects centroids probabilistically using feature relevance estimates derived from the data itself. Building on this, we propose two new feature selection algorithms, FS-MWK++, which aggregates feature weights across a range of Minkowski exponents identifying stable and informative features, and SFS-MWK++, a scalable variant based on subsampling. We support our approach with a theoretical analysis, demonstrating that, under explicit assumptions on noise features and cluster structure, relevant features are assigned consistently higher weights than noise features across a range of Minkowski exponents. Our software can be found at \url{https://github.com/xzhang4-ops1/FSMWK}.
\end{abstract}

\section{Introduction}
Clustering is a fundamental machine learning technique that assigns data points in a dataset to clusters, with each cluster containing similar data points. As clustering algorithms operate without the need for labelled data, they have been successfully applied across various domains, including data pre-processing, quantitative finance, image analysis, and bioinformatics \cite{ran2023comprehensive,oyewole2023data,zhou2021hiercc,wu2022construction}. However, their performance (and that of any other machine learning algorithm) can be heavily influenced by the quality and dimensionality of the data. High-dimensional data frequently contain redundant or irrelevant features, which may obscure meaningful patterns and degrade both efficiency and accuracy. Feature selection algorithms mitigate this issue by identifying and retaining only the most relevant features \cite{theng2024feature,solorio2022survey}. 

Unfortunately, most existing feature selection algorithms rely on labelled data, which may be noisy, corrupted, or entirely unavailable in many real-world scenarios. This limitation highlights the need for effective unsupervised feature selection methods, which can improve learning performance without requiring labels. Despite growing interest in unsupervised feature selection, scalable methods that perform well across diverse data geometries remain scarce. This is particularly limiting in domains such as genomics, anomaly detection, or remote sensing, where data is high-dimensional and labels are either unavailable or unreliable.

Clustering algorithms can be broadly categorised into different types, including density-based, hierarchical, fuzzy, and partitional methods \cite{gan2020data}. Density-based approaches, such as the classic DBSCAN \cite{schubert2017dbscan}, group points based on density regions, making them effective at detecting arbitrary-shaped clusters. Hierarchical algorithms construct a nested hierarchy of clusters, which can be visualised with a dendrogram. Fuzzy clustering algorithms assign each data point to every cluster with different degrees of membership, usually adding to one. Here, we focus on partitional clustering. 

Let \(X\) be a dataset, where each \(x_i \in X\) is described by \(m\) features. Here, \(i = 1, \ldots, n\) indexes data points, \(v = 1, \ldots, m\) indexes features, and \(l = 1, \ldots, k\) indexes clusters. Partitional algorithms produce a clustering $S=\{S_1, \ldots, S_k\}$ such that $X = \bigcup_{l=1}^k S_l$ and $S_l \cap S_t=\emptyset$ for $l,t = 1, \ldots, k$ with $l\neq t$. $k$-means \cite{macqueen1967some} is, arguably, the most popular algorithm for partitional clustering  \cite{jain2010data,ahmed2020k}. Like most clustering algorithms, $k$-means treats all features as equally contributing to cluster formation. However, in many real-world datasets, certain features are more informative than others. Hence, treating them equally can lead to poor clustering performance. 

Feature-weighted clustering methods address this limitation by assigning importance weights to features based on their contribution to the underlying clustering structure. Minkowski Weighted $k$-means (MWK) \cite{de2012minkowski} extends $k$-means by incorporating a feature weighting mechanism that adjusts the Minkowski distance metric based on the within-cluster dispersion of features. Although MWK improves cluster recovery, the computed feature weights are sensitive to the Minkowski exponent and initial centroids used (for details, see Section \ref{sec:related_work}). Our work addresses both of these issues by introducing a probabilistic, relevance-aware initialisation and using the stability of feature weights across exponents to guide unsupervised feature selection. In addition to empirical validation, we also provide a theoretical analysis supporting our proposed approach.

Our main contributions are threefold. First, we introduce MWK++, a novel initialisation strategy for Minkowski Weighted $k$-means that incorporates feature relevance estimates into the centroid selection process. Second, we propose FS-MWK++, a new method for unsupervised feature selection based on the stability of feature weights across a range of Minkowski exponents, which allows the identification of features that are consistently informative under different clustering geometries. Third, we propose SFS-MWK++, a scalable extension of FS-MWK++ based on subsampling, enabling efficient application to large datasets. These contributions are distinct from existing approaches, as they combine feature-weighted clustering with stability-based feature selection and scalable design.

\section{Related work}
\label{sec:related_work}

In this section, we describe the related work pertinent to this paper, focusing on two main areas. Section \ref{sec:related_work_clustering} presents a general overview of prominent clustering algorithms, while Section \ref{sec:related_work_fs} describes key unsupervised feature selection methods and associated challenges.

\subsection{Clustering}
\label{sec:related_work_clustering}

The $k$-means algorithm \cite{macqueen1967some} is a widely used partitional clustering method, which has been used and extended in numerous ways \cite{liu2023transforming,harris2022extensive,gao2023overview,ahmed2020k}. It produces a partition \(S=\{S_1, \ldots, S_k\}\) of a dataset $X$ such that \(S_l \cap S_t =\emptyset\) for all \(l,t=1, \ldots k\) with \(l\neq t\). This is achieved by minimising the objective function
\begin{equation}
W (S,Z) = \sum_{l=1}^{k} \sum_{x_i \in S_l} \sum_{v=1}^{m} (x_{iv} - z_{lv})^2,
\end{equation}
where $z_l \in Z$ is the centroid of cluster $S_l$. The minimisation follows three straightforward steps: (i) select $k$ data points from \(X\) uniformly at random, and set their values to \( z_1,...,z_k\); (ii) assign each \(x_i \in X\) to the cluster \(S_l\) represented by centroid \(z_l\) that is nearest to \(x_i\); (iii) update each \(z_l \in Z\) by setting each of its features to the mean of the same feature over the points in \(S_l\). If there are changes in \(Z\), go back to step (ii).

Although popular, $k$-means is not without drawbacks. For instance, it assumes that the features of \(X\) are equally relevant. This is problematic because it may lead to an irrelevant feature contributing to the objective just as much as a relevant feature. Moreover, even among relevant features there can be different degrees of relevance, and a robust algorithm should take this into account. Also, $k$-means is a greedy algorithm and by consequence it does not guarantee convergence to a global minimum. As a result, the choice of initial centroids is particularly important. Poor initialisation may lead to convergence to a suboptimal solution.

The $k$-means++ \cite{arthur2007k} algorithm addresses the latter of the problems above. It improves the initial centroids of $k$-means by spreading them out more strategically based on the data distribution. This often lowers the likelihood of poor local minima, usually leading to a better clustering \cite{ikotun2023k,han2022data}. The $k$-means++ algorithm is now the default $k$-means implementation in many popular software including MATLAB, scikit-learn, and R. Let \(d(x_i)\) be the distance from \(x_i\) to its nearest centroid in \(Z\). That is,
\[
d(x_i) = \min_{z_l \in Z} \sum_{v=1}^m (x_{iv}-z_{lv})^2,    
\]
$k$-means++ works by: (i) select a data point \(x_t \in X\) uniformly at random and set \(Z=\{x_t\}\);
(ii) select a new point \(x_t \in X\) with probability \(\frac{d(x_t)^2}{\sum_{i=1}^n d(x_i)^2}\), and set \(Z = Z \cup \{x_t\}\); (iii) repeat step (ii) until \(|Z|=k\).

The $k$-means and $k$-means++ algorithms rely on the Euclidean distance, making them biased towards Gaussian (spherical) clusters \cite{de2023k} and imposing the assumption that all features are equally relevant. The Minkowski weighted $k$-means (MWK) \cite{de2012minkowski} is a popular method (see for instance \cite{nino2021feature,deng2016survey,aradnia2022adaptive,melvin2016uncovering}) that overcomes these issues by introducing cluster-specific feature weights into the Minkowski distance. That is, the distance between a data point \(x_i\) and a centroid \(z_l\) is given by
\[
    d_p(x_i, z_l) = {\sum_{v = 1}^m} w_{lv}^p \lvert x_{iv} - z_{lv}\lvert^{p},
\]
where \(p > 1\) is the Minkowski exponent controlling the shape of the metric space. This leads to the objective function
\begin{equation}
\label{eq:mwk}
W_p (S,Z,w) = \sum_{l=1}^{k} \sum_{x_i \in S_l} \sum_{v =1}^{m} w_{lv}^{p} {\lvert x_{iv}-z_{lv} \rvert^p}.
\end{equation}

To minimise the above, MWK defines the dispersion of a feature \(v\) at a cluster \(S_l\) as \(D_{lv} = \sum_{x_i \in S_l} |x_{iv} - z_{lv}|^p\). By re-writing (\ref{eq:mwk}) as \(\sum_{l=1}^k \sum_{v=1}^m w_{lv}^p D_{lv}\) and minimising it subject to \(\sum_{v=1}^m w_{lv}=1\) for \(l=1, \ldots, k\), the following optimal weights are obtained
\begin{equation}
\label{eq:mwk_weights}
    w_{lv} = \frac{1}{\sum_{u=1}^m \left[\frac{D_{lv}}{D_{lu}}\right]^{\frac{1}{p-1}}}.
\end{equation}

In the above, \(w_{lv}\) represents the weight of feature \(v\) at cluster \(S_l\). For \(p>1\), this weight will be higher in features with values concentrated around the centroid (i.e., those with lower dispersion), thereby reflecting their degree of relevance. This aligns with the intuitive notion that a feature can have different degrees of importance at different clusters. Moreover, employing the Minkowski distance allows MWK to adapt to different cluster shapes, reducing the bias toward Gaussian clusters.

\subsection{Unsupervised feature selection}
\label{sec:related_work_fs}

High-dimensional data frequently contain redundant or irrelevant features, which can hide meaningful patterns, increase computational cost, and impair model generalisation. Feature selection addresses these challenges by identifying such features, thereby leading to more efficient, stable, and reliable results.

Feature selection using feature similarity (FSFS) \cite{mitra2002unsupervised} is a popular unsupervised feature selection method \cite{li2024exploring,kuzudisli2023review,yuan2021attribute}, which is also indicated by its high number of citations in Google Scholar. Hence, we include it in our comparison (see Section \ref{sec:results}). FSFS aims at detecting and discarding redundant features by calculating their maximum information compression index, 
\begin{align}
&2\lambda_2(v_i, v_j) = \mathrm{var}(v_i) + \mathrm{var}(v_j) \notag \\
&     - \sqrt{(\mathrm{var}(v_i) + \mathrm{var}(v_j))^2 - 4\,\mathrm{var}(v_i)\,\mathrm{var}(v_j)(1 - \rho(v_i, v_j)^2)} \label{eq:lambda2}
\end{align}
where \(\rho(v_i, v_j) \) is the Pearson correlation coefficient between features \(v_i\) and \(v_j\). Notice that \(\lambda_2\) measures the minimum information loss when projecting features to a lower dimension, hence, its use as a measure of information redundancy. FSFS proceeds as follows.

\begin{enumerate}
  \item Initialize the full feature set $V = \{1, \ldots, m\}$, and choose a parameter $\kappa$ such that $\kappa \leq m - 1$.
  \item Select the feature $v^* \in V$ with the lowest redundancy score $r^\kappa_{v}$. Retain $v^*$ and remove the $\kappa$ most similar features to $v^*$ from $V$.
  \item Set $\epsilon = r^\kappa_{v^*}$, and $\kappa = \min(\kappa, |V| - 1)$. If $\kappa = 1$, go to Step 6.
  \item While $r^k_v > \epsilon$, do the following:
  \begin{enumerate}
    \item Set $\kappa = \kappa - 1$.
    \item Set $r^\kappa_v = \min\limits_{v \in V} r^\kappa_v$1
    \item If $\kappa = 1$, go to Step 6.
  \end{enumerate}
  \item Go to Step 2.
  \item Return $V$.
\end{enumerate}
Notice that in FSFS, the parameter \(\kappa\) refers to the number of nearest neighbours considered. That is, \(r^\kappa_v\) is the dissimilarity between feature \(v\) and its \(\kappa^{th}\) nearest neighbour. Here, this use of \(\kappa\) is unrelated to the \(k\) of $k$-means.

Multi-Cluster Feature Selection (MCFS) \cite{cai2010unsupervised} is another popular unsupervised feature selection method \cite{theng2024feature,alhenawi2022feature}. MCFS combines spectral techniques from manifold learning with \(\ell_1\)-regularization to identify the most relevant features. The algorithm has a single parameter \(\kappa\), which specifies the number of neighbors in the \(\kappa\)-nearest neighbours graph. The original paper suggests setting \(\kappa=5\) as a default value. In our experiments (see Section \ref{sec:results}), we tuned this parameter from 1 to 5, and report the best results. While this procedure introduces a slight positive bias (i.e., it favours MCFS), it does not compromise our objectives. The algorithm is as follows.

\begin{enumerate}
  \item Generate a $\kappa$-nearest neighbours graph to model the local manifold structure of the data.
  \item Solve a generalised eigen-problem and retain the top  \(k\) eigenvectors, where \(k\) is the number of clusters.
  \item Solve $k$ $\ell_1$-regularised regression problems using Least Angle Regression (LAR) and obtain $k$ sparse coefficient vectors.
  \item For each feature $v = 1, \ldots, m$, compute its MCFS score.
  \item Output the features with the highest scores.
\end{enumerate}

We direct readers interested in further details to the original publication \cite{cai2010unsupervised}.

Spectral Feature Selection (SPEC) \cite{zhao2007spectral} is a unified framework for unsupervised and supervised feature selection based on spectral graph theory. It evaluates the relevance of each feature by measuring its consistency with the structure of a graph induced from pairwise instance similarities. Features are ranked according to how well they align with the leading eigenvectors of the normalised graph Laplacian, which capture the most coherent structural components of the data. We included SPEC as an established spectral baseline for unsupervised feature selection.

Unsupervised Discriminative Feature Selection (UDFS) \cite{yang2011} is a popular unsupervised feature selection method based on \(\ell_{2,1}\)-norm regularisation. UDFS aims to select features jointly, rather than one by one, by learning a transformation that exploits both discriminative information and feature correlations, while also preserving the local structure of the data. In this way, it differs from methods that rank features individually or rely only on the manifold structure induced by the full feature set. We included UDFS in our comparison as a well-established baseline for unsupervised feature selection.

Infinite Feature Selection (Inf-FS) \cite{roffo2020infinite} is an unsupervised feature selection method that evaluates feature importance by considering all possible feature subsets in a graph-based framework. Features are represented as nodes in a graph, with edges encoding pairwise relationships such as similarity or redundancy. The relevance of each feature is then computed by summing the contributions of all possible paths in the graph, effectively capturing both direct and indirect feature interactions. This formulation allows Inf-FS to rank features without requiring labels, while accounting for higher-order dependencies between them. We included Inf-FS as a non-deep-learning baseline that models global feature relationships.

Differentiable Unsupervised Feature Selection (DUFS) \cite{lindenbaum2021differentiable} is an unsupervised feature selection method based on a differentiable objective that combines a graph Laplacian-based score with a gating mechanism. DUFS learns continuously relaxed Bernoulli gates over the input features and optimises them through gradient-based training, allowing the graph Laplacian to be re-evaluated on gated subsets of features. In this way, the method aims to filter nuisance features while retaining those that best preserve the underlying structure of the data. We included DUFS as a strong baseline based on differentiable feature gating and Laplacian-driven feature selection.

Subspace Clustering Feature Selection (SCFS) \cite{parsa2020unsupervised} is a recent unsupervised feature selection method combining subspace clustering with sparse regression. SCFS aims to select features that preserve the multi-cluster structure of the data by adaptively learning sample similarities \cite{song2023review}. First, it computes a similarity matrix implicitly by solving a self-expressive model, in which each sample is represented through a low-dimensional space shared with other similar samples. Then, a regularised regression model is applied to link features to the discovered clustering structure. Features with stronger associations to the cluster are ranked higher. SCFS introduces two regularisation parameters, \(\alpha\) and \(\beta\), to control the trade-off between similarity preservation and sparsity. Following the original paper we tuned these parameters using grid search over the set \(\{10^{-4}, 10^{-2}, 1, 10^2, 10^4\}\) and selected the best performing combination. We direct readers interested in further details about SCFS to its original publication \cite{parsa2020unsupervised}.

Laplacian Score-regularized Concrete Autoencoder (LS-CAE) \cite{shaham2022deep} is a recent deep learning-based approach to unsupervised feature selection \cite{li2024exploring}. LS-CAE extends the Concrete Autoencoder (CAE) framework by incorporating a Laplacian score regularisation term into the loss function. This encourages the selection of features that both enable reconstruction of the original data and preserve the local clustering structure. To achieve this, LS-CAE trains a fully differentiable autoencoder, where feature selection is implemented through a Concrete layer that softly samples features in a learnable manner. The final selected features are obtained after annealing the sampling temperature towards discrete choices.

Although LS-CAE operates without class labels, and is therefore unsupervised, it differs from traditional unsupervised feature selection methods such as FSFS, MCFS, or SCFS. In particular, LS-CAE involves training a deep neural network optimised with multiple objectives, which arguably provides greater modelling capacity compared to classical methods based solely on feature scoring or graph-based analysis. As a result, LS-CAE may potentially have an advantage in certain settings. However, our experimental results (see Section \ref{sec:results}) demonstrate that other non-deep-learning methods (such as the one we introduce) can outperform LS-CAE, suggesting that deep architectures are not universally superior for unsupervised feature selection.

TabNet \cite{arik2021tabnet} is a deep learning architecture for tabular data that uses sequential attention to perform instance-wise feature selection. At each decision step, a learnable sparse mask is applied to the input features, directing the model's capacity towards the most salient ones. Feature importance 
scores are obtained by aggregating these masks across decision steps. In our experiments, TabNet was trained using ground-truth class labels, giving it a 
considerable advantage over the unsupervised methods in our comparison. We included it as a supervised deep learning baseline to assess whether our unsupervised approach can remain competitive even against methods with access to label information.

\section{Our proposed methods}

This section is divided into two parts. Section~\ref{sec:mwkpp} introduces a novel initialisation method for the Minkowski Weighted $k$-means (MWK) algorithm, which selects initial centroids by taking into account the relative relevance of features, thereby improving cluster recovery. Section~\ref{sec:new_fs} builds upon this foundation to develop a new unsupervised feature selection algorithm that analyses the stability of feature weights generated under varying parameter settings. Additionally, we propose a sampling-based extension that allows our method to scale effectively to high-dimensional data.

Each component of our approach represents a distinct contribution, namely a feature relevance-aware initialisation strategy, a stability-based feature selection framework, and a scalable variant for large datasets.

\subsection{The Minkowski weighted $k$-means++}
\label{sec:mwkpp}

Here, we propose a novel initialisation strategy for the MWK, taking inspiration from $k$-means++ (for details, see Section \ref{sec:related_work_clustering}). Our method, which we refer to as Minkowski Weighted $k$-means++ (MWK++) selects initial centroids by incorporating the relevance of features as estimated via their dispersions.

\begin{algorithm}[H]
    \caption{Minkowski Weighted $k$-means++ (MWK++)}
    \begin{algorithmic}[1]
        \Require Dataset \(X\), number of clusters \(k\), exponent \(p > 1\).
        \Ensure Initial centroids \(Z = \{z_1, \ldots, z_k\}\), initial feature weights \(w\).
        \State Select the first centroid \(z_1 \in X\) uniformly at random, and set \(Z = \{z_1\}\).
        \State Compute the Minkowski centre \(c \in \mathbb{R}^m\) of the dataset using exponent \(p\)
        \State For each feature \(v = 1, \ldots, m\) compute the dispersion
        \[
        D_v = \sum_{i=1}^n |x_{iv} - c_v|^p.
        \]
        \State Increment each \(D_v\) by the average of \(D\).
        \State Compute feature weights 
        \[
        %w_v = \left( \sum_{u=1}^m \left( \frac{D_v}{D_u} \right)^{1/(p-1)} \right)^{-1}.
        w_v = \frac{1}{\sum_{u=1}^m \left( \frac{D_v}{D_u} \right)^{\frac{1}{p-1}} }.
        \]
        \State Replicate \(w\) to form a \(k \times m\) matrix.
        \For{\(i = 2\) to \(k\)}
            \State For each \(x_i \in X\), compute distance to nearest centroid
            \[
            %d(x_i) = \min_{l=1, \ldots, k} \sum_{v=1}^m w_{lv}^p |x_{iv} - z_{lv}|^p.
            d(x_i) = \min_{z_l \in Z} \sum_{v=1}^m w_{lv}^p |x_{iv} - z_{lv}|^p.
            \]
            \State Set sampling probability
            \[
            P(x_i) = \frac{d(x_i)}{\sum_{j=1}^n d(x_j)}.
            \]
            \State Select one \(x_t \in X\) according to \(P(x_t)\), and add it to \(Z\).
        \EndFor
        \State \Return \(Z\), \(w\)
    \end{algorithmic}
\end{algorithm}

We note that, unlike the standard $k$-means++ initialisation which uses squared Euclidean distances, we employ the Minkowski distance directly in the sampling probability. This choice ensures consistency with the objective function of MWK and avoids introducing an additional exponent that would alter the relative scaling of distances. The addition of the average of \(D\) to each \(D_v\) acts as a regularisation term. In particular, it prevents degenerate cases where \(D_v = 0\), which would otherwise lead to undefined ratios in the computation of feature weights. While this operation slightly modifies the dispersion ratios, it improves numerical stability without affecting the overall behaviour of the method.

In the initialisation step, feature weights are computed globally and denoted by \(w_v\). These weights are then replicated across clusters to form \(w_{lv}\), which are subsequently updated during clustering. Our approach improves centroid selection by biasing the sampling process towards regions of the feature space where features have a lower dispersion, and are then more likely to be informative. By computing feature weights prior to sampling, MWK++ ensures that distances are measured with emphasis on relevant features. As a result, the selected centroids tend to be well-separated along the more informative dimensions, increasing the likelihood of high-quality clustering outcomes (for details, see Section \ref{sec:results}). In the next section, we show how feature weights computed with MWK++ can also serve as a foundation for our new unsupervised feature selection method.

We note that MWK++ modifies only the initialisation step of the MWK algorithm and does not alter its optimisation procedure. Since MWK follows an alternating minimisation scheme analogous to $k$-means, its convergence to a local minimum of the objective function is well established. Therefore, the use of MWK++ preserves these convergence properties.
\subsection{Feature Selection With MWK++}
\label{sec:new_fs}

This section introduces our Feature Selection method with MWK++ (FS-MWK++). This is a new unsupervised feature selection method based on the stability of feature weights computed by the MWK++ algorithm across a range of Minkowski exponents. Recall that the Minkowski distance induces different geometric biases depending on the value of \(p\). For example, in two dimensions, lower values such as \(p=1.1\) tend to favour diamond-shaped clusters, whereas \(p=2\) corresponds to the standard Euclidean distance and favours spherical clusters. By evaluating a broad range of \(p\) values, our method captures clustering structures under diverse distance biases, thereby reducing the risk of selecting features that are overly tailored to a single geometric bias.

Our proposed methods involve only a small number of parameters, none of which require tuning. The number of clusters \(k\) and the number of selected features \(r\) are provided as inputs. Rather than selecting a single Minkowski exponent, we consider a predefined set \(P\), allowing the method to capture different clustering geometries. Feature weights are computed automatically within the MWK framework based on within-cluster dispersions and updated iteratively during clustering. In the scalable variant, SFS-MWK++, the sample size \(n_s = k\sqrt{n}\) is fixed to balance representativeness and computational efficiency.

Rather than attempting to optimise \(p\) directly, we exploit the variability in clustering outcomes to assess the stability of feature relevance across multiple settings. For each \(p\), we run MWK++ multiple times and retain the feature weights corresponding to the lowest objective value. These weight vectors are then aggregated by taking the component-wise median, yielding a robust estimate of each feature's overall importance (for details, see Algorithm \ref{Alg:FS_MWKPP}). Features that consistently receive high weights across different clustering geometries are more likely to be genuinely informative. This stability-based approach naturally filters out noisy or unstable features without requiring supervision or parameter tuning.

More specifically, for a fixed exponent \(p\), MWK++ produces cluster-specific feature weights \(w_{lv}^{(p)}\), where \(l=1,\ldots,k\) and \(v=1,\ldots,m\). These weights are computed from the within-cluster dispersions
\[
D_{lv}^{(p)} = \sum_{x_i \in S_l} |x_{iv} - z_{lv}|^p,
\]
through
\[
w_{lv}^{(p)} = \frac{1}{\sum_{u=1}^m \left( \frac{D_{lv}^{(p)}}{D_{lu}^{(p)}} \right)^{\frac{1}{p-1}} }.
\]
For each value of \(p\), we run MWK++ multiple times and retain the weights corresponding to the clustering with the lowest objective value. This yields, for each \(p\), a \(k \times m\) matrix of retained weights. To obtain a single relevance score for each feature \(v\), we aggregate the corresponding weights over all pairs \((p,l)\), that is, over all exponents \(p \in P\) and all clusters \(l=1,\ldots,k\), and compute their median. Thus, the final score of feature \(v\) is the median of the set \(\{w_{lv}^{(p)} : p \in P,\; l=1,\ldots,k\}\). Finally, the features are ranked according to these aggregated scores, and the top \(r\) features are selected.

\begin{algorithm}[H]
    \caption{Feature Selection with MWK++ (FS-MWK++)}
    \label{Alg:FS_MWKPP}
    \begin{algorithmic}[1]
        \Require Dataset \(X\), number of clusters \(k\), number of features to select \(r\).
        \Ensure A subset containing the \(r\) most informative features.
        \State Define a set of Minkowski exponents \(P=\{1.1, \ldots, 3.0\}\).
        \ForAll{\(p \in P\)}
            \State Run MWK++ 25 times on \(X\), independently, with exponent \(p\).
            \State From these, identify the clustering that minimises ~\eqref{eq:mwk}.
            \State Retain the feature weights of the clustering identified in the previous step.
        \EndFor
        %\State Compute the component-wise median of the retained feature weights.
        \State For each feature \(v\), compute the median of \(w_{lv}^{(p)}\) over all retained pairs \((p,l)\).
        \State \Return The indices of the \(r\) features with the highest median weights.
    \end{algorithmic}
\end{algorithm}

Note that \(P\) denotes the set of Minkowski exponents considered, and \(p \in P\) is used in each individual run. To provide theoretical insight into feature selection, we identify a sufficient condition under which a feature consistently receives high weight across multiple clustering configurations. This condition provides the foundation for analysing when FS-MWK++ is expected to select a relevant feature.

\begin{dfn}
Let \( D_{lv}^{(p)} \) denote the within-cluster dispersion of feature \(v\) in cluster \(S_l\) computed with Minkowski exponent \(p\), and let \( w_{lv}^{(p)} \) denote the corresponding feature weight.
\end{dfn}

In the context of clustering, informative features are expected to reflect the underlying cluster structure, and therefore tend to exhibit relatively low within-cluster dispersion. In contrast, noise features do not align with this structure and are unlikely to consistently achieve low dispersion within clusters. This distinction motivates the following definition, which characterises noise features through their relative within-cluster dispersion.
\begin{dfn}
\label{def:noise}
A feature \(v\) is said to be a \emph{noise feature} if, for each cluster \(S_l\) and exponent \(p>1\),
\[
\frac{1}{|R|}\sum_{u \in R}
\left(\frac{D_{lv}^{(p)}}{D_{lu}^{(p)}}\right)^{\frac{1}{p-1}} > 1,
\]
where \(R\) denotes the set of relevant features. That is, on average, the dispersion of \(v\) exceeds that of relevant features when compared feature-wise.
\end{dfn}

\begin{lemma}
\label{lemma:low_w_noise_feature}
Let \(t\) be a noise feature as per Definition \ref{def:noise}, and suppose that all such features are drawn independently from the same distribution. If there exists at least one relevant feature, then for any \(p > 1\),
\[
w_{lt}^{(p)} < \frac{1}{m}.
\]
\end{lemma}
\begin{proof}
Let \(\mathcal{W}\) be the set of noise features. From Equation~\eqref{eq:mwk_weights}, we have that
\[
w_{lt}^{(p)}=
\frac{1}{\sum_{u=1}^m \left( \frac{D_{lt}^{(p)}}{D_{lu}^{(p)}} \right)^{\frac{1}{p-1}} }.
\]
Thus, it is enough to show that the denominator is greater than \(m\). We split the denominator into noise and relevant features:
\[
\sum_{u=1}^m \left( \frac{D_{lt}^{(p)}}{D_{lu}^{(p)}} \right)^{\frac{1}{p-1}}
=
\sum_{u \in \mathcal{W}} \left( \frac{D_{lt}^{(p)}}{D_{lu}^{(p)}} \right)^{\frac{1}{p-1}}
+
\sum_{u \notin \mathcal{W}} \left( \frac{D_{lt}^{(p)}}{D_{lu}^{(p)}} \right)^{\frac{1}{p-1}}.
\]

Since all noise features are drawn from the same distribution, their within-cluster dispersions are expected to be similar. Hence, for \(u \in \mathcal{W}\),
\[
\frac{D_{lt}^{(p)}}{D_{lu}^{(p)}} \approx 1,
\]
so
\[
\sum_{u \in \mathcal{W}} \left( \frac{D_{lt}^{(p)}}{D_{lu}^{(p)}} \right)^{\frac{1}{p-1}}
\approx |\mathcal{W}|.
\]

Now consider \(u \notin \mathcal{W}\), that is, relevant features. From Definition~\ref{def:noise}, we have
\[
\sum_{u \notin \mathcal{W}} \left( \frac{D_{lt}^{(p)}}{D_{lu}^{(p)}} \right)^{\frac{1}{p-1}} > m - |\mathcal{W}|.
\]

Combining the two parts, we obtain
\[
\sum_{u=1}^m \left( \frac{D_{lt}^{(p)}}{D_{lu}^{(p)}} \right)^{\frac{1}{p-1}} > m,
\]
and therefore
\[
w_{lt}^{(p)} < \frac{1}{m}.
\]
\end{proof}

\begin{lemma}
Let \(X\) contain both relevant and noise features. If each noise feature is drawn independently from a common distribution, then at least one feature will have a weight strictly higher than \(\frac{1}{m}\).    
\end{lemma}
\begin{proof}
From Lemma~\ref{lemma:low_w_noise_feature}, if \(t\) is a noise feature, then
\[
w_{lt}^{(p)} < \frac{1}{m}
\]
for any \(p>1\) and cluster \(S_l\). Now recall that, for each cluster \(S_l\), the feature weights satisfy
\[
\sum_{v=1}^m w_{lv}^{(p)} = 1.
\]
Therefore, if all features had weight at most \(\frac{1}{m}\), and all noise features had weight strictly smaller than \(\frac{1}{m}\), then the total sum of weights would be strictly smaller than 1, which is impossible. Hence, at least one of the remaining features must compensate for this deficit, and therefore must satisfy
\[
w_{lv}^{(p)} > \frac{1}{m}.
\]
This proves the result.
\end{proof}

The lemma above guarantees the existence of at least one feature with weight exceeding \(\frac{1}{m}\) in some cluster. This motivates the next lemma, which characterises the weight of such features at different values of \(p\). Since the dispersion ratios \(D_{lv}^{(p)}/D_{lu}^{(p)}\) vary smoothly with \(p\), they can be approximated by their values at \(p_0\) in a sufficiently small neighbourhood, making this a natural assumption.

\begin{lemma}
\label{lemma:delta}
Under the local approximation that, in a neighbourhood of \(p_0\), the ratios
\[
\frac{D_{lv}^{(p)}}{D_{lu}^{(p)}}
\]
can be approximated by
\[
\frac{D_{lv}^{(p_0)}}{D_{lu}^{(p_0)}}
\]
for all \(u=1,\ldots,m\). If \(w_{lv}^{(p_0)} \geq \frac{1}{m} + \gamma\) for some \(\gamma > 0\), then \(w_{lv}^{(p)} > \frac{1}{m}\) for all \(p \in (p_0 - \delta,\, p_0 + \delta)\), where 
\[
\delta = \gamma \cdot \frac{A(p_0)^2 (p_0 - 1)^2}{L(p_0)},
\]
with
\[
A(p_0) = \sum_{u=1}^m \left( \frac{D_{lv}^{(p_0)}}{D_{lu}^{(p_0)}} \right)^{1/(p_0 - 1)},
\]
and,
\[
L(p_0) = \sum_{u=1}^m \left( \frac{D_{lv}^{(p_0)}}{D_{lu}^{(p_0)}} \right)^{\frac{1}{p_0-1}} \left| \log \left( \frac{D_{lv}^{(p_0)}}{D_{lu}^{(p_0)}} \right) \right|.
\]
\end{lemma}

\begin{proof}
Let \(a_u = D_{lv}^{(p_0)} / D_{lu}^{(p_0)}\). Then
\[
w_{lv}^{(p)} =  \frac{1}{\sum_{u=1}^m \left[\frac{D_{lv}}{D_{lu}}\right]^{\frac{1}{p-1}}}\approx\frac{1}{\sum_{u=1}^m a_u^{1/(p - 1)}} = \frac{1}{A(p)}.
\]
Hence,
\[
\frac{d}{dp}w_{lv}^{(p)} = - \frac{A^\prime (p)}{A(p)^2}, \quad \text{where} \quad A^\prime (p) = -\sum_{u=1}^m \frac{a_u^{\frac{1}{p-1}} \log a_u}{(p-1)^2}.
\]
Define
\[
L(p) = \sum_{u=1}^m a_u^{\frac{1}{p-1}} |\log a_u|,
\]
so that
\[
|A^\prime(p)| \leq \frac{L(p)}{(p - 1)^2}, \quad \text{and} \quad |w_{lv}^\prime(p)| \leq \frac{L(p)}{A(p)^2 (p - 1)^2}.
\]

Let \(\delta\) be the minimum perturbation needed to decrease the weight of \(v\) to \(1/m\). Then by the Mean Value Theorem,
\[
\delta = \left(w_{lv}^{(p_0)} - \frac{1}{m} \right) \cdot \frac{A(p_0)^2 (p_0 - 1)^2}{L(p_0)}.
\]

Since \(w_{lv}^{(p_0)} \geq \frac{1}{m} + \gamma\), 
\[
\delta \geq \gamma \cdot \frac{A(p_0)^2 (p_0 - 1)^2}{L(p_0)},
\]
as claimed.
\end{proof}

Building on these lemmas, we now derive sufficient conditions under which a relevant feature is expected to be selected by FS-MWK++. These conditions link the smoothness of feature weights across \(p\), the proportion of clusters where the feature appears prominent, and the structure of the \(P\) grid. The result formalises the intuitive idea that stability and alignment with cluster structure make a feature likely to stand out.

\begin{theorem}
\label{thm:main}
Let \(X\) contain both relevant and noise features, and suppose noise features are drawn independently from a common distribution and are uncorrelated with cluster structure. Let \(v\) be a feature such that for a proportion \(\alpha\) of the \(k\) clusters, there exists a \(p_l \in P\) satisfying
\[
w_{lv}^{(p_l)} \geq \frac{1}{m} + \gamma,
\]
with \(\gamma>0\) and suppose that for all such \(p_l\),
\[
\gamma \cdot \frac{A(p_l)^2 (p_l - 1)^2}{L(p_l)}>\frac{1}{2\alpha}.
\]
Then FS-MWK++ is heuristically expected to select \(v\), provided that the number of selected features \(r\) is at least the number of features satisfying this condition.
\end{theorem}
\begin{proof}
Notice there are \(|P|\cdot k\) pairs \((p,l)\). To show that \(v\) is selected, it is sufficient to show that \(w_{lv}^{(p_l)}>\frac{1}{m}+\gamma\) for some \(\gamma>0\) in at least \(\frac{|P|\cdot k}{2}\) pairs. Let this weight condition hold for \(\alpha \cdot k\) clusters for each \(p_l\) in \(P\). Then,
\[
\alpha \cdot k \cdot |P|> \frac{|P|\cdot k}{2} \iff 2\alpha|P|>|P| \iff |P| > \frac{|P|}{2\alpha}.
\]
Recall that \(|P|=20\) and that the interval \([p - \delta,p + \delta]\) has length \(2\delta\). Given the step within \(P\) is of \(0.1\), we have 
\[
|P| \approx \frac{2\delta}{0.1}=20\delta=\delta\cdot |P| > \frac{|P|}{2\alpha} \iff \delta>\frac{1}{2\alpha}.
\]
Following Lemma \ref{lemma:delta}, the above leads to
\[
\gamma \cdot \frac{A(p_l)^2 (p_l - 1)^2}{L(p_l)}>\frac{1}{2\alpha}.
\]
\end{proof}

This result should be interpreted as a heuristic expectation, as it relies on a continuous approximation of the exponent grid and local smoothness assumptions.

Let us now demonstrate Theorem \ref{thm:main} with an example. Let \(m=12\), \(\gamma=0.15\), and \(\alpha=0.9\). That is, there exists a feature \(v\) such that for 90\% of the clusters \(S_l \in S\) there exists a (possibly different) \(p_l \in P\) for which \(w_{lv}^{(p_l)} \geq \frac{1}{m}+\gamma\). Recall that FS-MWK++ considers a total of \(|P| \cdot k\) pairs \((p,l)\). In this example, the feature is stated to have a high weight in only \(0.9 \cdot k\) pairs (as there is one \(p_l\) for each cluster). Hence, it is not immediately obvious FS-MWK++ would select \(v\) under mild assumptions. Notice that
\begin{align*}
&w_{lv}^{(p_l)} = \frac{1}{A(p_l)} \geq \frac{1}{m} + \gamma\\
&\Rightarrow A(p_l) \leq \frac{1}{\frac{1}{m}+\gamma} = \frac{1}{\frac{1}{12}+0.15} \approx 4.29. 
\end{align*}
    
Let us assume a value of approximately 80\%, \(A(p_l)=3.4\), and \(p_l=2.4\). Recall that \(a_u = D_{lv}^{(p_l)} / D_{lu}^{(p_l)}\), then
\[
A(p_l) =\sum_{u=1}^m a_u^{1/(p_l - 1)}= \sum_{u=1}^{12} a_u^{1/1.4}=3.4,
\]
leading to, on average, \(a_u^{1/1.4}\approx 0.283 \iff a_u \approx 0.171\). Let us use this to approximate \(L(p_l)\),
\[
L(p_l) = \sum_{u=1}^m a_u^{\frac{1}{p_l-1}} |\log a_u| \approx \sum_{u=1}^{12} 0.283|\log 0.171|\approx 6.00.
\]
Then,
\begin{align*}
&\gamma \cdot \frac{A(p_l)^2 (p_l - 1)^2}{L(p_l)}>\frac{1}{2\alpha}\\
&\iff 0.15 \frac{3.4^2 \cdot (2.4-1)^2}{6.00} \approx 0.566 >\frac{1}{2}.    
\end{align*}

Therefore, the condition in Theorem~\ref{thm:main} is satisfied, and \(v\) is expected to be selected by FS-MWK++.

The theoretical results above provide insight into how feature weights behave under the MWK framework. In particular, they show that noise features tend to receive weights below the uniform baseline \(\frac{1}{m}\), while relevant features are more likely to exceed this threshold and remain stable across nearby values of the Minkowski exponent \(p\). This motivates the design of FS-MWK++, which identifies informative features by aggregating their weights across multiple values of \(p\). In practice, for each exponent we retain the weight vector corresponding to the best clustering solution, and then compute the component-wise median across exponents. Features with consistently high weights are selected, reflecting both their relevance and stability under varying clustering geometries.

While FS-MWK++ is effective at identifying stable and informative features (for details, see Section~\ref{sec:results}), it becomes computationally expensive on large datasets. We address this limitation in three ways. First, we modify the computation of the Minkowski centre. Instead of solving an optimisation problem, we approximate the centre by using the component-wise median when \(p<1.5\), and the mean otherwise. This choice is motivated by robustness considerations: for lower values of \(p\), the Minkowski objective becomes more sensitive to deviations and noise, making the component-wise median a more stable estimator of central tendency. For larger values of \(p\), the objective behaves more smoothly, and the mean provides a more efficient and less variable estimate.

Second, we reduce the number of \(p\) values in our experiments by setting \(P=\{1.1, 1.3, 1.5,\ldots, 3.0\}\). That is, we have a set of 10 equally spaced values containing 1.1 and 3.0. Third, we introduce a scalable variant that further reduces the computational cost by operating on a representative subset of the data. This sampling-based alternative retains the core idea of stability across exponents while significantly improving runtime efficiency. Algorithm~\ref{Alg:SampleFS_MWKPP} describes the steps of this method. In our experiments, we fix the number of restarts and the set of Minkowski exponents \(P\) to balance robustness and computational efficiency. We observed that the results are stable for a range of reasonable values of these parameters, suggesting that the method is not overly sensitive to their precise choice.

\begin{algorithm}[H]
    \caption{Sample Feature Selection with MWK++ (SFS-MWK++)}
    \label{Alg:SampleFS_MWKPP}
    \begin{algorithmic}[1]
        \Require Dataset \(X \in \mathbb{R}^{nm}\), number of clusters \(k\), number of features \(r\).
        \Ensure A subset containing the \(r\) most informative features.
        \State Define a set of Minkowski exponents \(P=\{1.1, 1.3, 1.5, \ldots, 3.0\}\).
        \State Set the sample size \(n_s = k\sqrt{n}\).
        \For{i=1 to 25}
            \State Create a dataset \(X_s\) containing \(n_s\) data points from \(X\) selected uniformly at random.
            \ForAll{\(p \in P\)}
                \State Run MWK++ 25 times on \(X_s\), independently, with exponent \(p\).
                \State From these, identify the clustering that minimises ~\eqref{eq:mwk}.
                \State Retain the feature weights of the clustering identified in the previous step.
            \EndFor
        \EndFor
        %\State Compute the component-wise median of the retained feature weights.
        \State For each feature \(v\), compute the median of \(w_{lv}^{(p)}\) over all retained pairs \((p,l)\).
        \State \Return The indices of the \(r\) features with the highest median weights.
    \end{algorithmic}
\end{algorithm}

The sample size \(n_s = k\sqrt{n}\) is chosen to balance representativeness and computational efficiency. In particular, it ensures that, in expectation, each cluster is sufficiently represented in the subsample, while keeping the sample size sublinear in \(n\), which is essential for scalability to large datasets. Empirically, this choice provided a good trade-off between performance and runtime across all datasets considered.

We briefly discuss the computational complexity of the proposed methods. Let \(n\) be the number of data points, \(m\) the number of features, \(k\) the number of clusters, and \(I\) the number of iterations of the MWK algorithm. A single run of MWK has complexity \(\mathcal{O}(nmkI)\). In FS-MWK++, this procedure is repeated multiple times across different values of the Minkowski exponent and initialisations (25 restarts per \(p\)), leading to a computational cost that scales linearly with \(n\), as well as with the number of restarts and the size of the exponent set \(|P|\), but can be significant for large datasets. To address this, SFS-MWK++ operates on subsamples of size \(n_s = k\sqrt{n}\), reducing the per-run complexity to \(\mathcal{O}(n_s m k I)\). As a result, the overall complexity grows sublinearly with \(n\) within each clustering step, making the method more suitable for large-scale datasets.

\section{Experiments setting}

We divide our experiments into two main sets to separately evaluate (i) the clustering performance of the proposed MWK++ initialisation method, and (ii) the effectiveness of FS-MWK++ and SFS-MWK++ for unsupervised feature selection. The source code for our experiments can be found at \url{https://github.com/xzhang4-ops1/FSMWK}.

\subsection{Evaluating MWK++}

In our first set of experiments, we evaluate whether MWK++ offers improvements over the original Minkowski Weighted $k$-means (MWK), and also compare it against $k$-means++. The latter is the default $k$-means implementation in popular software packages, such as MATLAB, R, and scikit-learn. Hence, it has become a standard baseline in many widely used software packages. We compared these algorithms on synthetic datasets as these offer full control over the data generation process, enabling fair and reproducible comparisons under known ground truth. They also allow us to evaluate performance across a wide range of controlled cluster configurations, which would be unfeasible to obtain with real-world datasets.

We constructed 12 dataset configurations. For each of these configurations we generated 50 datasets, leading to a total of 600 datasets. Each of these datasets contains spherical Gaussian clusters, with each cluster defined by a diagonal covariance matrix whose variance, \(\sigma^2\), was sampled from a uniform distribution within \([0.5, 1.5]\). Hence, we have a mix of dense and sparse clusters. Cluster centroids were independently sampled from a multivariate normal distribution with zero mean and variance of one. The number of data points per cluster was sampled uniformly at random, with a minimum of 20 data points per cluster. To assess robustness under high-dimensional noise, we appended approximately 50\% additional noise features to each dataset. These noise features were sampled independently from a uniform distribution. 

For example, the dataset configuration 1000x4-5 +2NF contains 50 datasets with 1,000 data points originally described over 4 informative features and partitioned into 5 clusters, with two noise features added, resulting in 6 total features. We evaluated clustering performance using the Adjusted Rand Index (ARI) \cite{hubert1985comparing}, which quantifies the agreement between predicted and true labels while correcting for chance. 

\subsection{Evaluating FS-MWK++ and SFS-MWK++}

Our second set of experiments focuses on evaluating the effectiveness of our proposed feature selection methods, FS-MWK++ and SFS-MWK++, against well-established unsupervised feature selection baselines. We acknowledge that FS-MWK++ can be computationally expensive for large datasets (this is indeed the reason why we also introduced SFS-MWK++). Hence, we evaluate FS-MWK++ on our synthetic datasets as these are not too large. SFS-MWK++, on the other hand, was designed for scalability. Thus, we evaluate it on much larger real-world datasets we obtained from the popular UCI Machine Learning Repository \cite{kelly2025uci}. We have added approximately 10\% and 20\% noise features to each real-world dataset. For details, see Table \ref{tab:real_world_datasets}.

In the real-world datasets, it is perfectly possible that an original feature of the dataset is not actually relevant.
Given we do not have full knowledge of which features are truly relevant (or their degree of relevancy), we also evaluate performance using the Entropy of the resulting clusters (computed using ground-truth labels), under the assumption that lower entropy indicates a purer and more meaningful clustering. 

We compare our methods against MCFS, SCFS, FSFS, UDFS, Inf-FS, DUFS, LS-CAE, SPEC, and TabNet (for details, see Section \ref{sec:related_work_fs}). We normalised all datasets (real-world and synthetic), after adding the noise features, by the range. That is,
\[
x_{iv} = \frac{x_{iv} - \bar{x}_v}{\max (x_v) - \min (x_v)},
\]
where \(\bar{x}_v\), max(\(x_v\)), min(\(x_v\)) denote the mean, maximum, and minimum of feature \(v\), respectively. We selected the range normalisation instead of the more popular $z$-score because the former does not penalise features with multimodal distributions. This is an important consideration when clusters may be separable along such dimensions.

\begin{table}[H]\scriptsize
\caption{List of the real-world datasets used in our experiments. The column `Features' includes the noise features. We downloaded the datasets from the UCI machine learning repository \cite{kelly2025uci}, added uniformly random noise features, and then normalised by the range.}
\begin{center}
\begin{tabular}{l cccc  cccc  cc}
\toprule
&Data points&Clusters&Features&Noise\\
Dataset & $n$ & $k$& $m$ & Features\\
\midrule
CoverType +6NF  &581,012&7&60&6\\
CoverType +11NF &581,012&7&65&11\\
HandPostures +4NF&78,095&5&40&4\\
HandPostures +8NF&78,095&5&44&8\\
IDA2016 +17NF   &76,000&2&186&17\\
IDA2016 +34NF   &76,000&2&203&34\\
OnlineNewsPop +6NF &39,644&6&64&6\\
OnlineNewsPop +12NF&39,644&6&70&12\\
SkinSegmentation +1NF&245,057&2&4&1\\
SkinSegmentation +2NF&245,057&2&5&2\\
\bottomrule
\end{tabular}\\
\end{center}
\label{tab:real_world_datasets}
\end{table}

\section{Results and discussion}
\label{sec:results}
We present our experimental evaluation in two parts. First, we assess the clustering performance of MWK++ against standard baselines. Then, we evaluate the effectiveness of FS-MWK++ and its scalable variant, SFS-MWK++, in unsupervised feature selection tasks.

\subsection{Clustering performance of MWK++}

Table \ref{tab:mwkpp_results} reports the mean and standard deviation of the Adjusted Rand Index (ARI) across 50 datasets for each configuration. Each algorithm was run 25 times on each dataset, and the reported results reflect the average ARI across all runs. The table compares $k$-means++, MWK (using either the average across all values of the Minkowski exponent $p$, or the best value), and our proposed MWK++ under the same two evaluation schemes.

The results clearly demonstrate that MWK++ consistently outperforms both $k$-means++ and the original MWK across all configurations. Even when comparing MWK++ using all $p$ values (i.e., not selecting the best case), it achieves higher ARI than MWK in all configurations. When using the best exponent $p$, MWK++ yields the highest ARI in every case except one (2000x30-5 +15NF), where $k$-means++ slightly outperforms all variants. This exception is likely due to the relatively small number of clusters combined with a high number of informative features. In this setting, the effect of noise is limited, and the cluster structure can already be well captured by the Euclidean distance. As a result, the benefit of feature weighting is reduced, allowing $k$-means++ to perform competitively. This highlights that the advantage of MWK++ is most pronounced in settings with higher dimensionality, stronger noise, or more complex cluster structures.

These results confirm the benefits of our proposed initialisation strategy. The MWK++ method improves centroid selection by leveraging feature relevance estimates early in the clustering process, which leads to more informative separation of clusters. This advantage is especially evident in more challenging scenarios with high dimensionality or large numbers of clusters. For instance, in the configuration 2000x20-20 +10NF, the ARI improves from 0.45 (MWK, best $p$) to 0.66 with MWK++. In addition, the gap between MWK and MWK++ tends to widen as dimensionality and cluster complexity increase. Finally, the relatively low standard deviations across most configurations indicate that MWK++ provides not only better but also stable clustering results compared to its counterparts.

\begin{table*}[htbp]\scriptsize
\caption{Comparison in terms of mean Adjusted Rand Index between the clusterings obtained with $k$-means++, MWK, and MWK++. There are 50 datasets for each configuration. We run each algorithm 25 times on each dataset.}
\begin{center}
\begin{tabular}{l cccc  cccc  cc}
\toprule
&&&\multicolumn{4}{c}{MWK} & \multicolumn{4}{c}{MWK++} \\
 \cmidrule(lr){4-7} \cmidrule(lr){8-11} 
&\multicolumn{2}{c}{$k$means++}& \multicolumn{2}{c}{All $p$}&\multicolumn{2}{c}{Best $p$}& \multicolumn{2}{c}{All $p$}&\multicolumn{2}{c}{Best $p$}\\
\cmidrule(lr){2-3}\cmidrule(lr){4-5} \cmidrule(lr){6-7} \cmidrule(lr){8-9} \cmidrule(lr){10-11}
Dataset & Mean & Std& Mean & Std & Mean & Std & Mean & Std & Mean & Std\\
\midrule
1000x4-3 +2NF & 0.02 & 0.05& 0.17 & 0.06 & 0.28 & 0.06 & \textbf{0.31} & 0.06 & \textbf{0.44} & 0.05 \\
1000x4-5 +2NF& 0.05 & 0.01& 0.15 & 0.03 & 0.21 & 0.03 & \textbf{0.23} & 0.03 & \textbf{0.32} & 0.02 \\
1000x4-10  +2NF& 0.06 & 0.00& 0.08 & 0.01 & 0.11 & 0.01 & \textbf{0.11} & 0.01 & \textbf{0.17} & 0.01\\
1000x10-3 +5NF& 0.20 & 0.09& 0.59 & 0.05 & 0.83 & 0.06 & \textbf{0.71} & 0.04 & \textbf{0.87} & 0.06 \\
1000x10-5 +5NF& 0.06 & 0.03& 0.42 & 0.03 & 0.64 & 0.02 & \textbf{0.53} & 0.02 & \textbf{0.70} & 0.02 \\
1000x10-10 +5NF& 0.02 & 0.01& 0.21 & 0.21 & 0.33 & 0.02 & \textbf{0.39} & 0.01 & \textbf{0.51} & 0.01 \\
2000x20-5 +10NF& 0.54 & 0.06& 0.70 & 0.02 & 0.76 & 0.02 & \textbf{0.85} & 0.01 & \textbf{0.87} & 0.01 \\
2000x20-10 +10NF& 0.05 & 0.01& 0.53 & 0.01 & 0.64 & 0.01 & \textbf{0.72} & 0.01 & \textbf{0.78} & 0.02 \\
2000x20-20 +10NF& 0.02 & 0.01& 0.30 & 0.01 & 0.45 & 0.01 & \textbf{0.53} & 0.01 & \textbf{0.66} & 0.01 \\
2000x30-5 +15NF& \textbf{0.94} & 0.06& 0.77 & 0.02 & 0.81 & 0.01 & 0.87 & 0.02 & 0.88 & 0.03 \\
2000x30-10 +15NF& 0.39 & 0.06& 0.62 & 0.01 & 0.71 & 0.01 & \textbf{0.79} & 0.01 & \textbf{0.83} & 0.01 \\
2000x30-20 +15NF& 0.05 & 0.01& 0.46 & 0.01 & 0.61 & 0.01 & \textbf{0.69} & 0.02 & \textbf{0.77} & 0.01 \\
\bottomrule
\end{tabular}\\
\end{center}
\label{tab:mwkpp_results}
\end{table*}

\subsection{Unsupervised feature selection}

Let us first analyse FS-MWK++. Table~\ref{tab:fs_synthetic} presents the results for our FS-MWK++ experiments on synthetic datasets. For each method, we report the average proportion of correctly classified features(those that were either truly informative or correctly identified as noise), along with the corresponding standard deviation. Given these are synthetic datasets we know which features are composed solely of noise.

The results show that FS-MWK++ achieves outstanding performance across all dataset configurations, outperforming most baseline methods and achieving the best average performance overall. In fact, FS-MWK++ reaches perfect classification in 8 out of 12 configurations and achieves an average accuracy of 0.99 with a very low standard deviation (0.02), indicating both high precision and stability. This is a strong indication that feature weights derived from MWK++ are highly reliable indicators of feature relevance when aggregated across multiple exponents \(p\). LS-CAE remains a strong baseline and performs slightly better in the three configurations with only four informative features, likely due to its higher modelling capacity as a deep neural network. In contrast, FS-MWK++ does not rely on a learnable architecture or end-to-end training, yet still achieves superior overall performance. DUFS also performs competitively and matches FS-MWK++ in several cases. Overall, FS-MWK++ is the most accurate and consistent method on these synthetic datasets.

We also observe that recent unsupervised feature selection methods such as UDFS and DUFS perform competitively in several configurations, with DUFS in 
particular achieving strong and consistent performance. In contrast, Inf-FS and SPEC yield consistently poor results across all synthetic configurations. Upon inspection, we found that both methods return features in the opposite order to that intended, meaning the least important features were treated as the most important. This explains the near-identical accuracy of approximately 0.33 across all configurations: since noise features constitute roughly one third of the total features in our synthetic datasets, selecting them preferentially naturally yields results close to this value. However, as shown in Table~\ref{tab:fs_real_world}, Inf-FS performs more competitively on real-world datasets, indicating that it is better suited to more complex and naturally structured data. TabNet, despite having access to ground-truth labels during training, performs only moderately across synthetic configurations, and is consistently outperformed by FS-MWK++. This is a particularly noteworthy result, as it suggests that our fully unsupervised approach is able to identify informative features more reliably than a supervised deep learning method under these conditions. Overall, FS-MWK++ remains the most reliable method, maintaining near-perfect accuracy across all configurations.

These results also provide strong evidence that the proposed method is robust across a wide range of controlled scenarios. In particular, the synthetic datasets were constructed to systematically vary the number of clusters, the dimensionality, and the proportion of noise features. This allows us to evaluate the behaviour of FS-MWK++ under progressively more challenging conditions, including settings with high dimensionality and substantial noise. The consistently high performance across all configurations indicates that the method is not sensitive to a specific data geometry or parameter choice, but instead generalises well across different regimes. Furthermore, since the ground truth about informative and noise features is known in these datasets, the experiments provide a direct and reliable validation of the feature selection capability of the method. When combined with the results on real-world datasets, which demonstrate effectiveness in more complex and less controlled settings, we believe that the experimental evaluation provides a comprehensive validation of the proposed approach.

\begin{table*}[htbp]\scriptsize
\centering

\caption{Results for the feature selection experiments on synthetic datasets. We present the average proportion of correctly classified features (i.e., informative or non-informative), and related standard deviation.}
\resizebox{\textwidth}{!}{
\begin{tabular}{l cccccccccccccccccccc}
\toprule
&\multicolumn{2}{c}{MCFS}&
\multicolumn{2}{c}{SCFS}&
\multicolumn{2}{c}{FSFS}&
\multicolumn{2}{c}{LS-CAE}&
\multicolumn{2}{c}{UDFS}&
\multicolumn{2}{c}{Inf-FS}&
\multicolumn{2}{c}{DUFS}&
\multicolumn{2}{c}{SPEC}&
\multicolumn{2}{c}{TabNet}&
\multicolumn{2}{c}{FS-MWK++}\\
\cmidrule(lr){2-3} \cmidrule(lr){4-5} \cmidrule(lr){6-7}
\cmidrule(lr){8-9} \cmidrule(lr){10-11} \cmidrule(lr){12-13}
\cmidrule(lr){14-15} \cmidrule(lr){16-17} \cmidrule(lr){18-19}
\cmidrule(lr){20-21}
Dataset & Mean & Std & Mean & Std & Mean & Std & Mean & Std
& Mean & Std
& Mean & Std
& Mean & Std
& Mean & Std
& Mean & Std
& Mean & Std\\
\midrule

1000x4-3 +2NF  & 0.34 & 0.05 & 0.35 & 0.08 & 0.94 & 0.14 & \textbf{1.00} & 0.00 & 0.41 & 0.00 & 0.33 & 0.00 & 0.73 & 0.16 & 0.33 & 0.00 & 0.76 & 0.19 & 0.96 & 0.11\\
1000x4-5 +2NF  & 0.34 & 0.05 & 0.33 & 0.00 & 0.93 & 0.17 & \textbf{1.00} & 0.00 & 0.45 & 0.00 & 0.33 & 0.00 & 0.71 & 0.17 & 0.33 & 0.00 & 0.79 & 0.19 & 0.99 & 0.07\\
1000x4-10 +2NF & 0.41 & 0.17 & 0.33 & 0.00 & 0.97 & 0.12 & \textbf{1.00} & 0.00 & 0.48 & 0.00 & 0.33 & 0.00 & 0.67 & 0.18 & 0.33 & 0.00 & 0.84 & 0.19 & 0.98 & 0.08\\

1000x10-3 +5NF & 0.39 & 0.09 & 0.49 & 0.21 & 0.88 & 0.16 & 0.95 & 0.07 & 0.77 & 0.00 & 0.33 & 0.00 & \textbf{1.00} & 0.00 & 0.33 & 0.00 & 0.72 & 0.11 & \textbf{1.00} & 0.00\\
1000x10-5 +5NF & 0.36 & 0.06 & 0.35 & 0.06 & 0.88 & 0.13 & 0.92 & 0.07 & 0.79 & 0.00 & 0.33 & 0.00 & \textbf{1.00} & 0.00 & 0.33 & 0.00 & 0.78 & 0.10 & \textbf{1.00} & 0.00\\
1000x10-10 +5NF& 0.34 & 0.02 & 0.33 & 0.00 & 0.95 & 0.08 & 0.96 & 0.06 & 0.82 & 0.00 & 0.33 & 0.00 & \textbf{1.00} & 0.00 & 0.33 & 0.00 & 0.83 & 0.11 & \textbf{1.00} & 0.00\\

2000x20-5 +10NF  & 0.55 & 0.11 & 0.64 & 0.24 & 0.89 & 0.10 & 0.95 & 0.04 & 0.87 & 0.00 & 0.33 & 0.00 & 0.89 & 0.01 & 0.33 & 0.00 & 0.75 & 0.07 & \textbf{1.00} & 0.00\\
2000x20-10 +10NF & 0.38 & 0.06 & 0.33 & 0.00 & 0.93 & 0.08 & 0.96 & 0.03 & 0.91 & 0.00 & 0.33 & 0.00 & \textbf{1.00} & 0.01 & 0.33 & 0.00 & 0.79 & 0.07 & \textbf{1.00} & 0.00\\
2000x20-20 +10NF & 0.34 & 0.02 & 0.33 & 0.00 & 0.96 & 0.05 & 0.99 & 0.02 & 0.91 & 0.00 & 0.33 & 0.00 & 0.96 & 0.03 & 0.33 & 0.00 & 0.82 & 0.07 & \textbf{1.00} & 0.00\\

2000x30-5 +15NF  & 0.67 & 0.07 & 0.99 & 0.04 & 0.89 & 0.11 & 0.91 & 0.04 & 0.92 & 0.00 & 0.33 & 0.00 & 0.96 & 0.02 & 0.33 & 0.00 & 0.70 & 0.06 & \textbf{1.00} & 0.00\\
2000x30-10 +15NF & 0.62 & 0.11 & 0.50 & 0.24 & 0.93 & 0.07 & 0.96 & 0.02 & 0.92 & 0.00 & 0.33 & 0.00 & 0.94 & 0.03 & 0.33 & 0.00 & 0.73 & 0.06 & \textbf{1.00} & 0.00\\
2000x30-20 +15NF & 0.38 & 0.05 & 0.33 & 0.00 & 0.93 & 0.08 & 0.96 & 0.02 & 0.88 & 0.00 & 0.33 & 0.00 & 0.82 & 0.05 & 0.33 & 0.00 & 0.75 & 0.06 & \textbf{1.00} & 0.00\\

\hline
Average & 0.43 & 0.07 & 0.44 & 0.07 & 0.92 & 0.11 & 0.96 & 0.03
& 0.78 & 0.00
& 0.33 & 0.00
& 0.89 & 0.06
& 0.33 & 0.00
& 0.77 & 0.11
& \textbf{0.99} & 0.02 \\
\bottomrule
\end{tabular}
}
\label{tab:fs_synthetic}
\end{table*}

We now turn to the results of our scalable feature selection method, SFS-MWK++, evaluated on real-world datasets with added noise features. We cannot be certain we know which features are informative for each dataset. Hence, we measure performance with the proportion of selected features that were part of the original dataset (as with our previous experiments), and the entropy after feature selection. The latter measures the degree of class purity in each cluster. Lower entropy and higher proportion of correctly identified features both indicate better feature selection. Unfortunately, it was impractical to run SCFS, DUFS, and SPEC on these large datasets. SCFS and DUFS were excluded due to their computational and memory demands. SPEC was excluded for a more fundamental reason: it requires constructing and storing an $n \times n$ similarity matrix, which is infeasible for datasets of this scale. For instance, CoverType alone contains over 580,000 data points, making the required matrix too large to fit in memory.

SFS-MWK++ performs consistently well across all datasets. It achieves the highest or joint-highest proportion of original features selected in 9 out of 10 datasets, and simultaneously achieves the lowest entropy in the same cases. For example, in CoverType +11NF, SFS-MWK++ selects original features with 98\% accuracy and reduces entropy from 2.51 to 1.44, outperforming all competing methods. Similarly strong results are observed on IDA2016 +34NF and OnlineNewsPop +12NF, where SFS-MWK++ both avoids selecting noisy features and yields significantly purer clusters.

We also observe that the considered baselines exhibit different behaviours on real-world datasets. In particular, Inf-FS performs more competitively in this setting, achieving results comparable to classical methods such as FSFS and MCFS in several datasets. This contrasts with its behaviour on the synthetic benchmarks, where it yields nearly identical results across all configurations, suggesting that it is not well suited to the specific feature-selection scenario induced by our synthetic data generation procedure. UDFS shows moderate performance, but remains less competitive overall. TabNet, despite being trained with ground-truth labels, also performs poorly on the real-world datasets, remaining less competitive than SFS-MWK++ across all configurations. This further reinforces the 
effectiveness of our fully unsupervised approach in practical, large-scale settings.

These results highlight the effectiveness of SFS-MWK++ in preserving informative features under high-dimensional, noisy conditions. Despite its sampling-based approximation and lack of learning stage or supervision, SFS-MWK++ remains competitive with LS-CAE, which benefits from an end-to-end deep learning framework. Moreover, unlike LS-CAE and FSFS, which show signs of performance degradation in more complex datasets (e.g., IDA2016 +34NF and OnlineNewsPop +12NF), SFS-MWK++ maintains both low entropy and high proportion of original features selected.

Overall, the results in this section support both FS-MWK++ and SFS-MWK++ as effective unsupervised feature selection methods. FS-MWK++ demonstrates near-perfect accuracy in identifying informative and non-informative features in controlled synthetic settings, while SFS-MWK++ extends this success to real-world datasets, achieving strong noise rejection and improved clustering structure with minimal computational overhead. Together, they offer a practical and scalable solution for feature selection in both small and large-scale unsupervised learning tasks.

\begin{table*}[htbp]\scriptsize
 \centering
\caption{Comparison of feature selection methods on benchmark datasets. For each dataset, we report the original entropy \(H_{\text{orig}}\) (computed using ground-truth labels), and for each method, the proportion of selected features that were part of the original dataset, and the entropy after feature selection. A higher proportion suggests better discrimination against artificially added noise features. Lower entropy indicates purer clusters.}
%\resizebox{\textwidth}{!}{ 
%\setlength{\tabcolsep}{4pt}
%\renewcommand{\arraystretch}{1.2}
% \begin{tabular}{lccccccccc}
% \toprule
% & &\multicolumn{2}{c}{FSFS}&\multicolumn{2}{c}{MCFS}&\multicolumn{2}{c}{LS-CAE}&\multicolumn{2}{c}{SFS-MWK++}\\
% \cmidrule(lr){3-4} \cmidrule(lr){5-6} \cmidrule(lr){7-8} \cmidrule(lr){9-10}
% & \(H_{\text{orig}}\) & Prop. & H& Prop. & H& Prop. & H& Prop. & H\\
% \midrule 
% CoverType +11NF   & 2.51&0.91& 1.81 & 0.94& 1.54& 0.82& 2.11& \textbf{0.98}& \textbf{1.44}\\
% CoverType +6NF     & 2.05&0.93& 1.67& 0.93& 1.53& 0.92& 1.79 & \textbf{0.97}& \textbf{1.42}\\
% HandPostures +4NF  & 5.31& 0.95& 5.24& 0.90& 5.46& \textbf{1.00}& \textbf{5.02} & \textbf{1.00}& \textbf{5.02} \\
% HandPostures +8NF & 5.56& 0.95& 5.24& 0.70& 5.66& 0.95& 5.04 & \textbf{1.00}& \textbf{5.02}\\
% IDA2016 +17NF     & 1.94& 0.81& 2.12& 0.96& 1.58& 0.96& 1.49 & \textbf{0.98}& \textbf{1.35}\\
% IDA2016 +34NF     & 2.45& 0.66& 2.92& 0.87& 1.83& \textbf{0.97}& 1.46 & \textbf{0.97}& \textbf{1.38}\\
% OnlineNewsPop +12NF & 3.91& 0.66& 4.50& 0.76& 3.88& 0.80& 3.90& \textbf{0.86}& \textbf{3.54}\\
% OnlineNewsPop +6NF & 3.53& 0.84& 3.66& \textbf{0.91}& \textbf{3.27}& 0.88& 3.58& 0.87& 3.49\\
% SkinSegmentation +1NF& 7.67& 0.50& 7.70& 0.50& \textbf{7.56}& \textbf{1.00}& \textbf{7.56}& \textbf{1.00}& \textbf{7.56}\\
% SkinSegmentation +2NF & 7.74& 0.20& 7.87& 0.60& \textbf{7.56}& 0.60& 7.70 & \textbf{1.00} & \textbf{7.56}\\
% \bottomrule
% \end{tabular}
\resizebox{\textwidth}{!}{
\begin{tabular}{lccccccccccccccc}
\toprule
& &\multicolumn{2}{c}{FSFS}&
\multicolumn{2}{c}{MCFS}&
\multicolumn{2}{c}{LS-CAE}&
\multicolumn{2}{c}{Inf-FS}&
\multicolumn{2}{c}{UDFS}&
\multicolumn{2}{c}{TabNet}&
\multicolumn{2}{c}{SFS-MWK++}\\
\cmidrule(lr){3-4} \cmidrule(lr){5-6} \cmidrule(lr){7-8}
\cmidrule(lr){9-10} \cmidrule(lr){11-12} \cmidrule(lr){13-14} \cmidrule(lr){15-16}
& \(H_{\text{orig}}\) & Prop. & H & Prop. & H & Prop. & H 
& Prop. & H 
& Prop. & H 
& Prop. & H
& Prop. & H\\
\midrule 

CoverType +11NF   & 2.51 & 0.91 & 1.81 & 0.94 & 1.54 & 0.82 & 2.11 & 0.66 & 1.91 & 0.66 & 2.08 & 0.72 & 2.70 & \textbf{0.98} & \textbf{1.44}\\
CoverType +6NF    & 2.05 & 0.93 & 1.67 & 0.93 & 1.53 & 0.92 & 1.79 & 0.80 & 1.69 & 0.80 & 1.87 & 0.83 & 2.10 & \textbf{0.97} & \textbf{1.42}\\
HandPostures +4NF & 5.31 & 0.95 & 5.24 & 0.90 & 5.46 & \textbf{1.00} & \textbf{5.02} & 0.80 & 5.12 & 0.80 & 5.16 & 0.85 & 5.31 & \textbf{1.00} & \textbf{5.02}\\
HandPostures +8NF & 5.56 & 0.95 & 5.24 & 0.70 & 5.66 & 0.95 & 5.04 & 0.64 & 5.42 & 0.64 & 5.16 & 0.72 & 5.68 & \textbf{1.00} & \textbf{5.02}\\
IDA2016 +17NF     & 1.94 & 0.81 & 2.12 & 0.96 & 1.58 & 0.96 & 1.49 & 0.82 & 1.95 & 0.82 & 2.07 & 0.82 & 2.04 & \textbf{0.98} & \textbf{1.35}\\
IDA2016 +34NF     & 2.45 & 0.66 & 2.92 & 0.87 & 1.83 & \textbf{0.97} & 1.46 & 0.67 & 2.49 & 0.67 & 2.85 & 0.67 & 2.70 & \textbf{0.97} & \textbf{1.38}\\
OnlineNewsPop +12NF & 3.91 & 0.66 & 4.50 & 0.76 & 3.88 & 0.80 & 3.90 & 0.66 & 3.88 & 0.66 & 3.57 & 0.73 & 3.79 & \textbf{0.86} & \textbf{3.54}\\
OnlineNewsPop +6NF & 3.53 & 0.84 & 3.66 & \textbf{0.91} & \textbf{3.27} & 0.88 & 3.58 & 0.81 & 3.56 & 0.81 & 3.20 & 0.84 & 3.54 & 0.87 & 3.49\\
SkinSegmentation +1NF & 7.67 & 0.50 & 7.70 & 0.50 & \textbf{7.56} & \textbf{1.00} & \textbf{7.56} & 0.50 & 7.73 & 0.50 & 7.56 & 0.90 & 7.60 & \textbf{1.00} & \textbf{7.56}\\
SkinSegmentation +2NF & 7.74 & 0.20 & 7.87 & 0.60 & \textbf{7.56} & 0.60 & 7.70 & 0.20 & 7.87 & 0.20 & 7.56 & 0.68 & 7.68 & \textbf{1.00} & \textbf{7.56}\\

\bottomrule
\end{tabular}
}
\label{tab:fs_real_world}
\end{table*}

Although LS-CAE benefits from a deep learning architecture, it requires training a neural network with multiple optimisation steps, which is computationally demanding. In contrast, SFS-MWK++ relies on repeated clustering on subsamples and does not involve gradient-based training. This makes it significantly more scalable in practice, particularly for large datasets, while still achieving competitive or superior performance.

\section{Conclusion}

This paper introduces a novel initialisation method for the Minkowski Weighted $k$-means (MWK) algorithm, we called the Minkowski Weighted $k$-means++ (MWK++). We then used this as a foundational step to design two new unsupervised feature selection algorithms, FS-MWK++ and SFS-MWK++. Our contributions are motivated by the limitations of existing clustering and feature selection methods in high-dimensional, label-free settings, where noise can severely degrade performance. In addition to extensive empirical evaluation, we provide a theoretical analysis that supports the effectiveness of our feature selection strategy under mild and realistic assumptions.

We showed that MWK++ consistently outperforms both $k$-means++ and the original MWK across a wide variety of synthetic data configurations, especially in high-dimensional or noisy settings. By incorporating feature relevance into the initial centroid selection process, MWK++ improves both clustering accuracy and stability, without introducing significant computational overhead. 

Building on this, we proposed FS-MWK++, a feature selection method based on aggregating feature weights across multiple distance exponents. This approach avoids the need for label supervision or deep learning infrastructure while achieving near-perfect discrimination between informative and noisy features in synthetic datasets. To address scalability, we introduced SFS-MWK++, a sampling-based extension that retains the effectiveness of FS-MWK++ while significantly reducing computational cost. 

Experiments on real-world datasets with added noise confirmed that SFS-MWK++ matches or outperforms state-of-the-art baselines, including the deep learning-based LS-CAE and TabNet, the latter of which had access to ground-truth labels during training. Overall, our results highlight the potential of clustering-driven feature weighting as a robust foundation for unsupervised learning tasks. Future work will explore theoretical guarantees, and extend our methods to more general clustering frameworks beyond MWK++.

In future work, we plan to investigate adaptive strategies for selecting the range of Minkowski exponents, with the goal of further improving performance and robustness. In addition, we will explore the application of the proposed approach to more complex data types, such as structured or temporal data. Finally, we aim to strengthen the theoretical analysis by relaxing some of the assumptions and deriving general guarantees.

\bibliographystyle{ieeetr}
\bibliography{references}

\end{document}